\documentclass[letterpaper]{article}
\usepackage{uai2020}
\usepackage[margin=1in]{geometry}

\usepackage{bm}
\usepackage{amsthm}
\usepackage{amsmath}
\usepackage{amsfonts}
\usepackage{natbib}
\usepackage[colorlinks,citecolor=blue,urlcolor=blue,filecolor=blue,backref=page]{hyperref}
\usepackage{graphicx}

\usepackage{amsmath}

\usepackage{algpseudocode}
\usepackage{algorithm}
\usepackage{enumitem}
\usepackage{graphicx}
\usepackage{dirtytalk}
\usepackage{color}

\usepackage{amsmath}
\usepackage{amsthm}

\usepackage{subcaption}

\DeclareCaptionLabelFormat{andtable}{#1~#2  \&  \tablename~\thetable}

\DeclareMathOperator*{\argmax}{arg\,max}

\newtheorem{proposition}{Proposition}

\usepackage{times}

\title{Stochastic Gradient MCMC with Repulsive Forces}

\author{ {\bf Victor Gallego} \\
Institute of Mathematical Sciences (ICMAT) \&\\
SAMSI, Duke University \\
\url{victor.gallego@icmat.es}
\And
{\bf David Rios Insua}  \\
Institute of Mathematical Sciences (ICMAT) \& \\
SAMSI, Duke University \\
\url{david.rios@icmat.es}
}

\begin{document}

\maketitle

\begin{abstract}
We propose a unifying view of two different Bayesian inference algorithms, Stochastic Gradient Markov Chain Monte Carlo (SG-MCMC) and Stein Variational Gradient Descent (SVGD), leading to improved and efficient novel sampling schemes. We show that SVGD combined with a noise term can be framed as a multiple chain SG-MCMC method. Instead of treating each parallel chain independently from others, our proposed algorithm implements a repulsive force between particles, avoiding collapse and facilitating a better exploration of the parameter space. We also show how the addition of this noise term is necessary to obtain a valid SG-MCMC sampler, a significant difference with SVGD. 
Experiments with both synthetic distributions and real datasets illustrate the benefits of the proposed scheme.
\end{abstract}

\section{INTRODUCTION}

Bayesian computation lies at the heart of many machine learning models in both academia and industry, \cite{bishop2006pattern}. Thus, it is of major importance to develop efficient approximation techniques that tackle the intractable integrals that arise in large scale Bayesian inference and prediction problems, \cite{gelman2013bayesian}. 

Recent developments in Bayesian techniques applied to large scale datasets and deep models include variational based approaches such as \emph{Automatic Differentiation Variational Inference} (ADVI), \cite{blei2017variational}, and \emph{Stein Variational Gradient Descent} (SVGD),  \cite{liu2016stein}, or sampling approaches such as \emph{Stochastic Gradient Markov Chain Monte Carlo} (SG-MCMC), \cite{NIPS2015_5891}. While variational techniques enjoy faster computations, they rely on optimizing a family of posterior approximates that may not contain the actual posterior distribution, potentially leading to severe bias and underestimation of uncertainty, \cite{pmlr-v80-yao18a} or \cite{48127}. 

There has been recent interest in bridging the gap between variational Bayes and MCMC techniques, see e.g. \cite{zhang2018}, to develop new scalable approaches for Bayesian inference, e.g., \cite{carbonetto2012}.
In this work, we draw on a similitude between the SG-MCMC and SVGD approaches to propose a novel family of very efficient sampling algorithm. 
Any competing MCMC approach should verify the following list of properties as our proposal will do:
\begin{itemize}
    \item \emph{scalability}. For this, we resort to SG-MCMC methods since at each iteration they may be approximated to just require a minibatch of the dataset,
    \item \emph{convergence to the actual posterior}, and
    \item \emph{flexibility}. Since we provide a parametric formulation of the transition kernel, it is possible to adapt other methods such as \emph{Hamiltonian Monte Carlo}, \cite{chen2014stochastic}, or the Nos\'e-Hoover thermostat method, \cite{ding2014bayesian}.
\end{itemize}

Our contributions are summarized as follows. First, we provide a unifying hybrid scheme of SG-MCMC and SVGD algorithms satisfying the previous list of requirements. Second, and based on the previous connection, 
we develop new SG-MCMC schema that include repulsive forces between particles. We also study the behaviour of SVGD and our new sampler via their respective Fokker-Planck equations, showing a significant difference which makes our proposed sampler a true SG-MCMC sampler. 

\begin{figure}[!h]
    \centering
\includegraphics[width=0.45\textwidth]{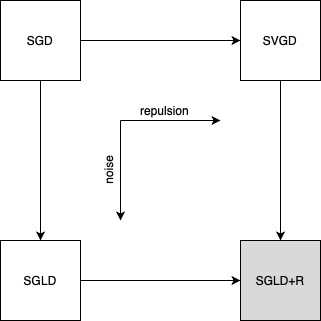}
    \caption{Relationships between different samplers. In light gray, our proposed sampler.}\label{fig:diagram}
\end{figure}

Figure \ref{fig:diagram} provides a diagram of our contributed scheme. Starting with SGD, one can add a carefully crafted noise term to arrive at the simplest SG-MCMC sampler, SGLD. Then, one can add repulsion between particles to speed-up the mixing time, leading to SGLD+R, our main contribution (Section \ref{sec:framework}). 

After an overview of posterior approximation methods in Section \ref{sec:back}, we propose our framework in Section \ref{sec:framework}.
Section \ref{sec:experiments} discusses relevant experiments showcasing the benefits of our proposal. Finally, Section \ref{sec:conclusion} sums up our contributions and highlight several open problems.

\section{BACKGROUND AND RELATED WORK}\label{sec:back}
Consider a probabilistic model $p(\bm{x}| \bm{z})$ and a prior distribution $p(\bm{z})$ where $\bm{x}$ denotes an observation and $\bm{z} = ( z_1, \ldots, z_d) $ an unobserved $d-$dimensional latent variable or parameter, depending on the context. We are interested in performing inference regarding the unobserved variable $\bm{z}$, by approximating its posterior distribution:
$$
p(\bm{z} | \bm{x}) = \frac{ p(\bm{z})p(\bm{x}| \bm{z}) }{ \int p(\bm{z})p(\bm{x}| \bm{z}) d\bm{z} } = \frac{ p(\bm{z})p(\bm{x}| \bm{z}) }{ p(\bm{x}) } = \frac{p(\bm{z},\bm{x})}{p(\bm{x})}.
$$
Except for reduced classes of distributions like \textit{conjugate priors}, \cite{raiffa1961applied}, the previous integral $p(\bm{x}) = \int p(\bm{z})p(\bm{x}| \bm{z}) d\bm{z}$ is analytically intractable; no general explicit expressions of the posterior are available. Thus, several techniques have been proposed to perform approximate posterior inference.

\subsection{Inference as sampling}\label{sec:infassamp}

Hamiltonian Monte Carlo (HMC, \cite{neal2011mcmc}) is an effective sampling method for models whose probability is point-wise computable and differentiable. 
HMC requires the exact simulation of a certain dynamical system which can be cumbersome in high-dimensional or large data settings. When this is an issue, \cite{welling2011bayesian} proposed a formulation of a continuous-time Markov process that converges to a target distribution $p(\bm{z} | \bm{x})$. It is based on the Euler-Maruyama discretization of Langevin dynamics:
\begin{eqnarray}\label{eq:sgld}
\bm{z}_{t+1} \leftarrow \bm{z}_{t} - \epsilon_t \nabla \log p(\bm{z}_t,\bm{x})  + \mathcal{N}(\bm{0}, 2\epsilon_t I),
\end{eqnarray}
where $\epsilon_t$ is the step size. The previous iteration uses the gradient evaluated at one data point $\bm{x}$, but we can use the full dataset or mini-batches. Several extensions of the original Langevin sampler have been proposed to increase mixing speed, see for instance \cite{li2016preconditioned,li2016high,li2019communication}.

\cite{NIPS2015_5891} proposed a general formulation of a continuous-time Markov process that converges to a target distribution $\pi(\bm{z}) \propto \exp (-H(\bm{z}))$. It is based on the Euler-Maruyama discretization of the generalized Langevin dynamics:
\begin{align}\label{eq:sgmcmc}
\begin{split}
&\bm{z}_{t+1} \leftarrow \bm{z}_{t} \\ &- \epsilon_t \left[ (\mathbf{D}(\bm{z}_t) + \mathbf{Q}(\bm{z}_t)) \nabla H(\bm{z}_t) + \mathbf{\Gamma}(\bm{z}_t) \right] + \mathcal{N}(\mathbf{0}, 2\epsilon_t \mathbf{D}(\bm{z}_t)),
\end{split}
\end{align}
with $\mathbf{D}(\bm{z})$ being a diffusion matrix; $\mathbf{Q}(\bm{z})$, a curl matrix and $\mathbf{\Gamma}(\bm{z})_i = \sum_{j=1}^d \frac{\partial}{\partial \bm{z}_j} (\mathbf{D}_{ij}(\bm{z}) + \mathbf{Q}_{ij}(\bm{z})) $ is a correction term which amends the bias.

To obtain a valid SG-MCMC algorithm, we simply have to choose the dimensionality of $\bm{z}$ (e.g., if we augment the space with auxiliary variables as in HMC), and the matrices $\mathbf{D}$ and $\mathbf{Q}$. For instance, the popular Stochastic Gradient Langevin Dynamics (SGLD) algorithm, the first SG-MCMC scheme in \cite{welling2011bayesian} in (\ref{eq:sgld}), is obtained when $\mathbf{D} = \mathbf{I}$ and $\mathbf{Q} = \mathbf{0}$. In addition, the Hamiltonian variant can be recovered if we augment the state space with a $d-$dimensional momentum term $\mathbf{m}$, leading to an augmented latent space $ \bar{\bm{z}} = (\bm{z}, \mathbf{m})$. Then, we set $\mathbf{D} = \mathbf{0}$ and $\mathbf{Q} = \begin{pmatrix}
\mathbf{0} & -\mathbf{I} \\
\mathbf{I} & \mathbf{0}
\end{pmatrix}$.

\subsection{Inference as optimization}\label{sec:iasopt}


Variational inference, \cite{kucukelbir2017automatic}, tackles the problem of approximating the posterior $p(\bm{z} | \bm{x})$ with a tractable parameterized distribution $q_{\bm{\lambda}}(\bm{z}|\bm{x})$. The goal is to find parameters $\bm{\lambda}$ so that the variational distribution $q_{\bm{\lambda}}(\bm{z}|\bm{x})$ (also referred to as the variational guide or variational approximation) is as close as possible to the actual posterior. Closeness is typically measured through 
Kullback-Leibler 
divergence $KL(q || p)$, which is reformulated into the \emph{evidence lower bound} (ELBO), given by
\begin{equation}\label{eq:elbo}
\mbox{ELBO}(q) = \mathbb{E}_{q_{\bm{\lambda}}(\bm{z}|\bm{x})} \left[ \log p(\bm{x},\bm{z}) - \log q_{\bm{\lambda}}(\bm{z}|\bm{x})\right].
\end{equation}
To allow 
for greater flexibility, typically a deep non-linear model conditioned on observation $\bm{x}$ defines the mean $\mu_{\bm{\lambda}}(\bm{x})$ and covariance matrix $\sigma_{\bm{\lambda}}(\bm{x})$ of a
Gaussian distribution $q_{\bm{\lambda}}(\bm{z}|\bm{x}) \sim \mathcal{N}(\mu_{\bm{\lambda}}(\bm{x}), \sigma_{\bm{\lambda}}(\bm{x}))$.
Inference is then performed using a gradient-based maximization routine, leading to variational parameter updates
\begin{align*}
\bm{\lambda}_{t+1} = \bm{\lambda}_t +  \epsilon_t \nabla_{\bm{\lambda}} \mbox{ELBO}(q),
\end{align*}
where $\epsilon$ is the learning rate. Stochastic gradient descent (SGD), \cite{hoffman2013stochastic}, or some variant of it, such as Adam, \cite{kingma2014adam}, are used as optimization algorithms.

On the other hand, SVGD, \cite{liu2016stein}, frames posterior sampling as an optimization process, in which a set of $L$ particles $\lbrace \bm{z}_i \rbrace_{i=1}^L$ is evolved iteratively via a velocity field $\bm{z}_{i, t+1} \leftarrow \bm{z}_{i, t} + \epsilon_t \phi(\bm{z}_{i,t})$, where $\phi : \mathbb{R}^d \rightarrow \mathbb{R}^d$ is a smooth function characterizing the perturbation of the latent space.
Assume that $q$ is the particle distribution at iteration $t$ and $q_{\left[ \epsilon \phi \right]}$ the distribution after update ($t+1$). Then, the optimal choice of the velocity field $\phi$ can be framed through the optimization problem $\phi^* = \argmax_{\phi \in \mathcal{F}} \lbrace -\frac{d}{d\epsilon} \mbox{KL}( q_{\left[ \epsilon \phi \right]} \| p) \rbrace$, in which $\phi$ is chosen to maximize the decreasing rate on the Kullback-Leibler (KL) divergence between the particle distribution and the target, and $\mathcal{F}$ is some proper function space. When $\mathcal{F}$ is a \emph{reproducing kernel Hilbert space} (RKHS), \cite{liu2016stein} showed that the optimal velocity field leads to
\begin{align}\label{eq:svgd}
\begin{split}
&\bm{z}_{i, t+1} \leftarrow \bm{z}_{i, t} \\&- \epsilon_t \frac{1}{L}  \sum_{j=1}^L \left[ k(\bm{z}_{j, t}, \bm{z}_{i, t})\nabla H(\bm{z}_{j,t}) + \nabla_{\bm{z}_{j, t}}  k(\bm{z}_{j, t}, \bm{z}_{i, t})  \right],
\end{split}
\end{align}
where the RBF kernel $k(\bm{z}, \bm{z}') = \exp (-\frac{1}{h}  \| \bm{z} - \bm{z}' \|^2 )$ is typically adopted. Observe that in \ref{eq:svgd} the gradient term $\nabla_{\bm{z}_{j, t}}  k(\bm{z}_{j, t}, \bm{z}_{i, t})$ acts as a repulsive force that prevents particles from collapsing. 



\subsection{The Fokker-Planck equation}\label{sec:fp}
Consider a SDE of the form $d\bm{z} = \mu_t(\bm{z})dt + \sqrt{2 D_t(\bm{z})}dB_t$. The distribution $q_t(\bm{z})$ of a population of particles evolving according to the previous SDE from some initial distribution $q_0(\bm{z})$ is governed by the Fokker-Planck partial differential equation (PDE), \cite{risken-fpe-1989},
$$
\frac{\partial}{\partial t} q_t(\bm{z}) = -\frac{\partial}{\partial \bm{z}} \left[ \mu_t(\bm{z}) q_t(\bm{z})\right] + \frac{\partial^2}{\partial \bm{z}^2} \left[ D_t(\bm{z})q_t(\bm{z})\right].
$$
Deriving the Fokker-Planck equation from the SDE of a potential SG-MCMC sampler is of great interest since we can check whether the target distribution is a stationary solution of the previous PDE (and thus, the sampler is consistent); and to compare if two a priori different SG-MCMC samplers result in the same trajectories.

\subsection{Related work}

There has been recent interest in developing new dynamics for SG-MCMC samplers with the aim of exploring the target distribution more efficiently. \cite{chen2014stochastic} proposed a stochastic gradient version of HMC, and \cite{ding2014bayesian} did the same leveraging for the Nos\'e-Hoover thermostat dynamics. \cite{chen2016bridging} leverages ideas from stochastic gradient optimization by proposing the analogue sampler to the Adam optimizer. A relativistic variant of Hamiltonian dynamics was introduced by \cite{abbati2018adageo}. Two of the most recent derivations of SG-MCMC samplers are \cite{zhang2019cyclical}, in which the authors propose a policy controlling the learning rate which serves as a better preconditioning; and \cite{gong2019meta}, in which a meta-learning approach is proposed to learn an efficient SG-MCMC transition kernel.

While the previous works propose new kernels which can empirically work well, they focus on the case of a single chain. We instead consider the case of several chains in parallel, and the bulk of our contribution focuses on how to develop transition kernels which can exploit interactions between chains. Thus, our framework can be seen as an orthogonal development to the previous listed approaches, and could be combined with them.

On the theoretical side, \cite{chen2018unified} study another connection between MCMC and deterministic flows, in particular they explore the correspondence between Langevin dynamics and Wasserstein gradient flows.
We instead formulate the dynamics of SVGD as a particular kind of MCMC dynamics, which enables us to use the Fokker-Planck equation to show in an straightforward manner how our samplers are valid. \cite{liu2017stein} started to consider similitudes between SG-MCMC and SVGD, though in this work we propose the first hybrid scheme between both methods.

\section{THE PROPOSED FRAMEWORK}\label{sec:framework}

We use the framework from \cite{NIPS2015_5891} in an augmented state space $\bm{z} = \left( \bm{z}_{1}, \bm{z}_{2}, \ldots,\bm{z}_{L}\right) \in \mathbb{R}^{Ld}$ to obtain a valid posterior sampler that runs multiple ($L$) Markov chains with interactions. This version of SG-MCMC is given by the equation
\begin{equation}\label{eq:general}
\bm{z}_{t+1} \leftarrow \bm{z}_t -\epsilon_t \left[ (\mathbf{\mathbf{D_K}} + \mathbf{Q_K})\mathbf{\nabla} + \mathbf{\Gamma_K} \right] + \bm{\eta}_t,
\end{equation}
with $\bm{\eta}_t \sim \mathcal{N}(\mathbf{0}, 2\epsilon_t \mathbf{\mathbf{D_K}})$. Now, $\bm{z}_t = \left(\bm{z}_{1,t} \ldots  \bm{z}_{L,t} \right)^\top$ is an $Ld$-dimensional vector defined by the concatenation of $L$ particles; $\mathbf{\nabla} \in \mathbb{R}^{L \times d \times 1}$ so that $(\mathbf{\nabla})_{i,:} = \nabla H(\bm{z}_{i,t})$\footnote{Though $\mathbf{\nabla} \in \mathbb{R}^{L d \times 1}$ to allow multiplication by $\mathbf{\mathbf{D_K}} + \mathbf{Q_K}$, we reshape it as $\mathbf{\nabla} \in \mathbb{R}^{L\times d \times 1}$ to better illustrate how it is defined.}; $\mathbf{\mathbf{D_K}} \in \mathbb{R}^{Ld\times Ld}$ is an expansion of the diffusion matrix $D$, accounting for the distance between particles; $\mathbf{Q_K} \in \mathbb{R}^{Ld\times Ld}$ is the curl matrix, which is skew-symmetric and might be used if a Hamiltonian variant is adopted; and $\mathbf{\Gamma_K}$ is the correction term from the framework of \cite{NIPS2015_5891}. Note that $\mathbf{\mathbf{D_K}}$, $\mathbf{Q_K}$ and $\mathbf{\Gamma_K}$ can depend on the state $\bm{z}_t$ (an example will be given below), but we do not make it explicit to simplify notation.

In matrix form, the update rule (\ref{eq:svgd}) for SVGD can be expressed as
\begin{equation}\label{eq:svgd_mat}
\overline{\bm{z}}_{t+1} \leftarrow \overline{\bm{z}}_t -\frac{\epsilon_t}{L}\left( \overline{\mathbf{K}} \overline{\mathbf{\nabla}} + \overline{\mathbf{\Gamma}} \right)
\end{equation}
where $\overline{\mathbf{K}} \in \mathbb{R}^{L \times L}$ so that $(\overline{\mathbf{K}})_{ij} = k(\bm{z}_i, \bm{z}_j)$, $\overline{\mathbf{\nabla}} \in \mathbb{R}^{L\times d}$ and $\overline{\bm{z}}_t \in \mathbb{R}^{L\times d}$. Casting the later matrix as a tensor $\mathbf{\nabla} \in \mathbb{R}^{L \times d \times 1}$ and the former one as a tensor $\mathbf{K} \in \mathbb{R}^{(L \times d) \times (L \times d)}$ by broadcasting along the second and fourth axes, we may associate $\mathbf{K}$ with the SG-MCMC's diffusion matrix $\mathbf{D}$ over a $Ld-$dimensional space. 

The big matrix $\mathbf{\mathbf{D_K}}$ in Eq. (\ref{eq:general}) is defined as a permuted block-diagonal matrix consisting of $d$ repeated kernel matrices $\overline{\mathbf{K}}$:
$$
\mathbf{\mathbf{D_K}} = \left[
\begin{array}{cccc}
\overline{\mathbf{K}} &  &  &  \\
 & \overline{\mathbf{K}} &  &  \\
 &  & \ddots &   \\
 &  &  & \overline{\mathbf{K}}  \\
\end{array}
\right] \mathbf{P},
$$
with $\mathbf{P}$ being the $Ld \times Ld$ permutation matrix
$$
\mathbf{P} = \left[
\begin{array}{c|c|c|c}
\begin{matrix}
1 & & &\\
 & & &\\
 & & &\\
  & & &\\
\end{matrix} &\begin{matrix}
 & & & \\
1 & & &\\
 & & & \\
  & & & \\
\end{matrix}  & \ddots & \begin{matrix}
 & & &\\
 & & &\\
 & & &\\
1 & & &\\
\end{matrix} \\
\hline
\begin{matrix}
 &1 & &\\
 & & &\\
 & & &\\
  & & &\\
\end{matrix} &\begin{matrix}
 & & & \\
 & 1& &\\
 & & & \\
  & & & \\
\end{matrix}  & \ddots & \begin{matrix}
 & & &\\
 & & &\\
 & & &\\
 &1 & &\\
\end{matrix} \\
\hline
\ddots &\ddots & \ddots & \ddots \\
\hline
\begin{matrix}
 & & &1\\
 & & &\\
 & & &\\
  & & &\\
\end{matrix} &\begin{matrix}
 & & & \\
 & & &1\\
 & & & \\
  & & & \\
\end{matrix}  & \ddots & \begin{matrix}
 & & &\\
 & & &\\
 & & &\\
& & &1\\
\end{matrix} \\
\end{array}
\right].
$$
The permutation matrix $\mathbf{P}$ rearranges the block-diagonal kernel matrix to match with the dimension ordering of the state space $\bm{z}_t = \left(\bm{z}_{1,t} \ldots  \bm{z}_{L,t} \right)^\top$.
With this convention, $\mathbf{\mathbf{D_K}} \mathbf{\nabla}$ is equivalent to $\overline{\mathbf{K}} \overline{\mathbf{\nabla}}$, only differing in the shape of the resulting matrix. This allows us to frame SVGD plus the noise term as a valid scheme within the SG-MCMC framework of \cite{NIPS2015_5891}, using the $Ld$-dimensional augmented state space.



From this perspective, Eq. \eqref{eq:svgd_mat} can be seen as a special case of Eq. \eqref{eq:general} with curl matrix $\mathbf{Q_K} = \mathbf{0}$ and no noise term. We refer to this perturbed variant of SVGD as \emph{Parallel SGLD plus repulsion} (SGLD+R):
\begin{equation}\label{eq:psvgd_mat}
\bm{z}_{t+1} \leftarrow \bm{z}_t -\frac{\epsilon_t}{L}\left( \mathbf{\mathbf{D_K}} \mathbf{\nabla} + \mathbf{\Gamma_K} \right) + \bm{\eta}_t, \quad \bm{\eta}_t \sim \mathcal{N}(\mathbf{0}, 2\epsilon_t \mathbf{\mathbf{D_K}}/L).
\end{equation}
Since $\mathbf{\mathbf{D_K}}$ is a definite positive matrix (constructed from the RBF kernel), we may use the key result from \cite{NIPS2015_5891} (Theorem 1) to derive the following:

\begin{proposition}
SGLD+R (or its general form, Eq. \eqref{eq:general}) has $\pi(\bm{z}) = \prod_{l=1}^L \pi(\bm{z}_l)$ as stationary distribution, and the proposed discretizations are asymptotically exact as $\epsilon_t \rightarrow 0$.
\end{proposition}
\noindent Having shown that SVGD plus a noise term can be framed as an SG-MCMC method, we may now explore the design spaces of the $\mathbf{\mathbf{D_K}}$ and $\mathbf{Q_K}$ matrices. 

Algorithm \ref{alg:alg1} shows how to set the sampler up. The step sizes $\epsilon_t$ decrease to $0$ using the Robbins-Monro conditions, \cite{robbins1951stochastic}, given by $\sum_{t=1}^\infty \epsilon_t = \infty, \sum_{t=1}^\infty \epsilon_t^2 < \infty$. However, in practical situations we can consider a small and constant step size, see Section \ref{sec:experiments}. 

\begin{algorithm}[!h] %
\caption{Bayesian Inference via SGLD+R}  
\label{alg:alg1}
\begin{algorithmic}
\State {\bf Input:} A target distribution with density function $\pi(\bm{z}) \propto \exp (-H(\bm{z}))$ and a prior distribution $p(\bm{z})$. 
\State {\bf Output:} A set of particles $\{\bm{z}_i\}_{i=1}^{ML}$ that approximates the target distribution.  
\State Sample initial set of particles from prior: $\bm{z}_1^0, \bm{z}_2^0, \ldots \bm{z}_L^0 \sim p(\bm{z})$.
\For{each iteration $t$}
\vspace{-1.5\baselineskip}
\State 
\begin{align} \label{eq:psvgd_alg}
\begin{split}
&\bm{z}_i^{t+1}  \gets  \bm{z}_i^t \\& - \epsilon_t \frac{1}{L}\sum_{l=1}^L\big[  k(\bm{z}_l^t, \bm{z}_i^t)  \nabla_{\bm{z}_l^t} H(\bm{z}_l^t) + \nabla_{\bm{z}_l^t} k(\bm{z}_l^t, \bm{z}_i^t)\big] + \bm{\eta}_i^t
\end{split}
\end{align}
\State where $\bm{\eta}_i^t$ is the noise for particle $i$ defined as in Eq (\ref{eq:psvgd_mat}).
\State After  burn-in period, start collecting particles: $ \{\bm{z}_i\}_{i=1}^{NL} \gets \{\bm{z}_i\}_{i=1}^{(N-1)L} \cup \{ \bm{z}_1^{t+1}, \ldots,  \bm{z}_L^{t+1} \} $ 
\EndFor
\end{algorithmic}
\end{algorithm}

\paragraph{Complexity.}  Finally, our proposed method is amenable to sub-sampling, since the mini-batch setting from SG-MCMC can be adopted: the main computational bottleneck lies in the evaluation of the gradient $\nabla_{\bm{z}} \log \pi( \bm{z})$, which can be troublesome in a big data setting such as when $\pi(\bm{z} | \mathcal{D}) \propto p(\bm{z}) \Pi_{i=1}^N p(\bm{x}_i | \bm{z})  $. We may then  approximate the true gradient with an unbiased estimator taken along a minibatch of datapoints $\Omega \subset \lbrace 1, 2, \ldots, N \rbrace$ in the usual way:
$$
\nabla_{\bm{z}} \log \pi( \bm{z} | \mathcal{D}) \approx \nabla_{\bm{z}} \log p(\bm{z}) + \frac{N}{| \Omega |} \sum_{i \in \Omega} \nabla_{\bm{z}} \log p(\bm{x}_i | \bm{z}).
$$

As with the original SVGD algorithm, the complexity of the update rule (\ref{eq:svgd_mat}) is $\mathcal{O}(L^2)$, with $L$ being the number of particles, since we need to evaluate kernels of signature $k(z_i, z_j)$. Using current state-of-the-art automatic differentiation frameworks, such as \texttt{jax}, \cite{jax2018github}, we can straightforwardly compile kernels using \emph{just-in-time} compilation, \cite{frostig2018compiling}, at the cost of a negligible overhead compared to parallel SGLD for moderate values of $L \sim 50$ particles.

If many more particles are to be used, one could approximate the expectation in Eq. (\ref{eq:svgd_mat}) using subsampling at each iteration, as proposed by the authors of SVGD, or by using more sophisticated approaches from the molecular dynamics literature, such as the Barnes-Hutt algorithm, \cite{barnes1986hierarchical}, to arrive at an efficient $\mathcal{O}(L \log L)$ computational burden at a negligible approximation error. We leave this approximation for further work.

\subsection{Relationship with SVGD}\label{sec:relationship}

In this Section we study in detail the behaviour of SVGD and SGLD+R. To do so, we will derive the Fokker-Planck equation for the SGLD+R sampler first.
\begin{proposition}
The distribution $q_t(\bm{z})$ of a population of particles evolving according to SGLD+R is governed by
$$
\frac{\partial}{\partial t} q_t(\bm{z}) = -\frac{\partial}{\partial \bm{z}} \left[ (\mathbf{D_K} \nabla \log \pi(\bm{z}) + \mathbf{\Gamma_K}) q_t(\bm{z})\right] + \frac{\partial^2}{\partial \bm{z}^2} \left[ \mathbf{D_K} q_t(\bm{z})  \right].
$$
The target distribution $\pi(\bm{z})$ is a stationary solution of the previous PDE.
\end{proposition}
\begin{proof}
The first part is a straightforward application of the Fokker-Planck equation from Section \ref{sec:fp}. For the last part, we need to show that
\begin{align*}
\frac{\partial}{\partial t} q_t(\bm{z}) = 0 = &-\frac{\partial}{\partial \bm{z}} \left[ (\mathbf{D_K} \nabla \log \pi(\bm{z}) + \mathbf{\Gamma_K}) \pi(\bm{z})\right] + \\& \frac{\partial^2}{\partial \bm{z}^2} \left[ \mathbf{D_K} \pi(\bm{z})  \right].
\end{align*}
To see so, we will expand each term in the rhs:
\begin{align*}
& \frac{\partial}{\partial \bm{z}} \left[ (\mathbf{D_K} \nabla \log \pi(\bm{z}) + \mathbf{\Gamma_K}) \pi(\bm{z})\right] = \\
&= \pi(\bm{z})\frac{\partial}{\partial \bm{z}} \left[ \mathbf{D_K} \nabla \log \pi(\bm{z}) + \mathbf{\Gamma_K} \right] + \\& \left[ \mathbf{D_K} \nabla \log \pi(\bm{z}) + \mathbf{\Gamma_K} \right] \frac{\partial}{\partial \bm{z}} \pi(\bm{z}) = \\
&= \pi(\bm{z})\frac{\partial}{\partial \bm{z}}\mathbf{D_K} \nabla \log \pi(\bm{z})+ \pi(\bm{z})\mathbf{D_K} \nabla^2 \log \pi(\bm{z}) + \\& \pi(\bm{z}) \frac{\partial}{\partial \bm{z}} \mathbf{\Gamma_K} + \mathbf{D_K} \nabla \log \pi(\bm{z}) \nabla \pi(\bm{z})  + \mathbf{\Gamma_K} \nabla \pi(\bm{z}) = \\
&= \mathbf{\Gamma_K} \nabla \pi(\bm{z}) + \pi(\bm{z})\frac{\partial}{\partial \bm{z}}\mathbf{D_K}\nabla \log \pi(\bm{z}) +  \\&
\pi(\bm{z}) \frac{\partial}{\partial \bm{z}} \mathbf{\Gamma_K} +
\mathbf{D_K}(\pi(\bm{z})\nabla^2 \log \pi(\bm{z}) + \nabla \log \pi(\bm{z}) \nabla \pi(\bm{z})).
\end{align*}
The other term expands to
\begin{align*}
&\frac{\partial^2}{\partial \bm{z}^2} \left[ \mathbf{D_K} \pi(\bm{z})  \right] = \\
&= \frac{\partial}{\partial \bm{z}} \left[ \frac{\partial}{\partial \bm{z}} \mathbf{D_K} \pi(\bm{z}) + \mathbf{D_K} \nabla \pi(\bm{z}) \right] = \\
&= \frac{\partial^2}{\partial \bm{z}^2}\mathbf{D_K} \pi(\bm{z}) + \pi(\bm{z}) \nabla \log \pi(\bm{z}) \frac{\partial}{\partial \bm{z}}\mathbf{D_K} + \\ &\frac{\partial}{\partial \bm{z}}  \mathbf{D_K} \nabla \pi(\bm{z}) + \mathbf{D_K} \nabla^2 \pi(\bm{z}) = \\
&= \frac{\partial}{\partial \bm{z}}  \mathbf{D_K} \nabla \pi(\bm{z}) +
\pi(\bm{z}) \nabla \log \pi(\bm{z}) \frac{\partial}{\partial \bm{z}}\mathbf{D_K} + \\
&\frac{\partial^2}{\partial \bm{z}^2}\mathbf{D_K} \pi(\bm{z})+ 
\mathbf{D_K}(\pi(\bm{z})\nabla^2 \log \pi(\bm{z}) + \nabla \log \pi(\bm{z}) \nabla \pi(\bm{z})).
\end{align*}
Taking into account that by the definition of the correction term, $\mathbf{\Gamma_K} = \frac{\partial}{\partial \bm{z}} \mathbf{D_K}$, the previous expansions are equal so they cancel each other in the rhs of the PDE.
\end{proof}
This last result is an alternative proof of our Preposition 1, without having to resort to the framework of \cite{NIPS2015_5891} as was the case there. It is of independent interest for us here, since we can establish a complementary result for the case of SVGD as follows.
\begin{proposition}
The distribution $q_t(\bm{z})$ of a population of particles evolving according to SVGD is governed by
$$
\frac{\partial}{\partial t} q_t(\bm{z}) = -\frac{\partial}{\partial \bm{z}} \left[ (\mathbf{D_K} \nabla \log \pi(\bm{z}) + \mathbf{\Gamma_K}) q_t(\bm{z})\right] .
$$
The target distribution $\pi(\bm{z})$ is not a stationary solution of the previous PDE in general.
\end{proposition}
\begin{proof}
As before, the first part is a straightforward application of the Fokker-Planck equation from Section \ref{sec:fp}. For the last part, note that the difference with Proposition 2 is that the term $\frac{\partial^2}{\partial \bm{z}^2} \left[ \mathbf{D_K} q_t(\bm{z})  \right]$ is absent now in the PDE, which prevents $\pi(\bm{z})$ from being a stationary solution in general.
\end{proof}
The term $\dfrac{\partial^2}{\partial \bm{z}^2} \left[ \mathbf{D_K} q_t(\bm{z})  \right]$ encourages the entropy in the distribution $q_t(\bm{z})$. By ignoring it, the SVGD flow achieves stationary solutions that underestimate the variance of the target distribution. On the other hand, SGLD+R, performs a correction, leading to the correct target distribution. In the next example we highlight this fact in a relatively simple setting.

\begin{figure}[!ht]
    \centering
    \includegraphics[width=5cm]{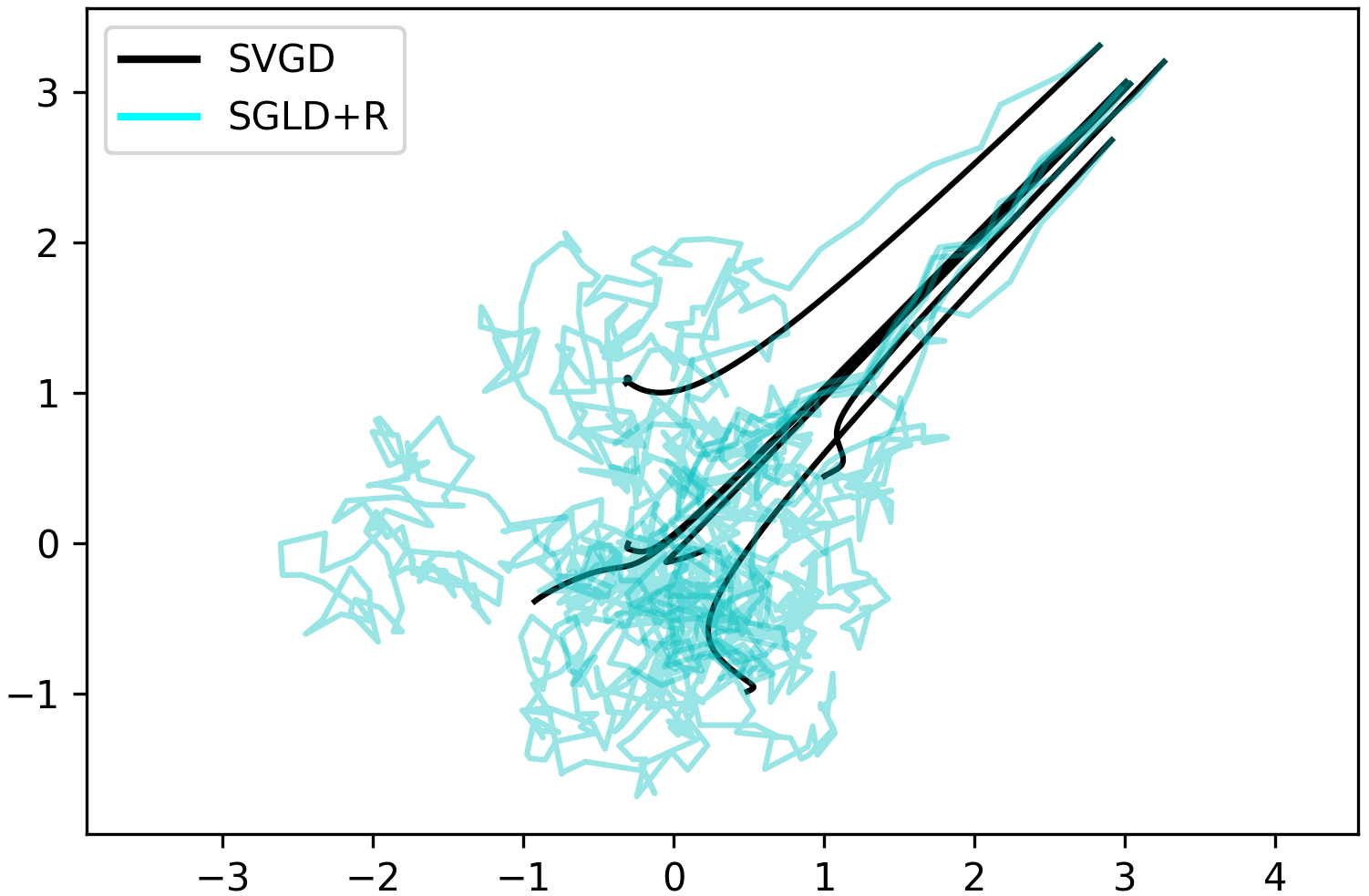}
    \qquad
    \begin{tabular}[b]{cccc}\hline
      \textbf{Estimates} & \textbf{SVGD} & \textbf{SGLD+R} & \textbf{Target} \\ \hline
      $\mu$ &  $0.18$ &$\textbf{0.08}$  & 0.00\\
      $\sigma_{x}$ & $0.70$ & $\textbf{0.90}$ & 1.00\\
      $\sigma_{y}$  & $0.74$ & $\textbf{0.87}$ & 1.00\\
    \hline
    \end{tabular}
    \captionlistentry[table]{A table beside a figure}
    \captionsetup{labelformat=andtable}
    \caption{Trajectories of the compared samplers. The table depicts estimates of quantities of interest for the standard Gaussian target.}
    \label{fig:comp}
  \end{figure}

\paragraph{Example.} Consider a standard bi-dimensional Gaussian target, $\pi(\bm{z}) \sim \mathcal{N}(0, I)$. The initial distribution of particles is $p(\bm{z}) = q_0(\bm{z}) \sim \mathcal{N}([3,3], \mbox{diag}([0.25, 0.25]))$. We let both samplers run for $T = 200$ iterations using $L = 6$ particles, and we plot their trajectories in Figure \ref{fig:comp}. Note how SGLD+R, since it is a valid sampler, explores a greater region of the target distribution, in comparison with SVGD, which underestimates the extension of the actual target. This phenomenon was predicted by Propositions 2 and 3. We also attach a Table, in which we report estimates of the target mean, $\mu$, and marginal standard deviations, $\sigma_x$ and $\sigma_y$, respectively. Notice how SGLD+R estimates are closer to the ground truth values for the standard Gaussian target of this example.

\section{EXPERIMENTS}\label{sec:experiments}

This Section describes the experiments developed to empirically test the proposed scheme. After the example from Section \ref{sec:relationship}, which compares SVGD and SGLD+R, we focus on confronting SGLD+R with the non-repulsive variant. First, we deal with two synthetic distributions, which offer a moderate account of complexity in the form of multimodality. In our second group of experiments, the focus is on a more challenging setting, testing a deep Bayesian model over several benchmark real data sets. 

Code for the different samplers is open sourced at \url{https://github.com/anon3232/sgmcmc-force}. We rely on the library \texttt{jax}, \cite{jax2018github}, as the main package, since it provides convenient automatic differentiation features with \emph{just-in-time} compilation which is extremely useful in our case for an efficient implementation of the SG-MCMC transition kernels.

\subsection{Synthetic distributions} 
The goal of this experiment is to see how well the samples generated by our framework approximate some quantities of interest, which can be analytically computed since the distributions are known. We thus test our proposed scheme with the following distributions.

\begin{figure}[!htb]
    \centering
    \begin{subfigure}[b]{0.2\textwidth}
        \includegraphics[width=\textwidth]{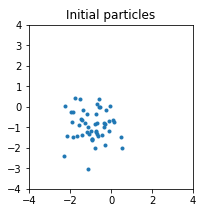}
        \caption{Prior particles}
        \label{fig:gull}
    \end{subfigure}
    \begin{subfigure}[b]{0.2\textwidth}
        \includegraphics[width=\textwidth]{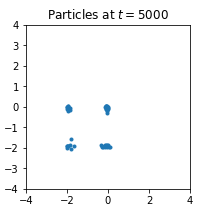}
        \caption{SGLD dyn.}
        \label{fig:tiger}
    \end{subfigure}
    \begin{subfigure}[b]{0.2\textwidth}
        \includegraphics[width=\textwidth]{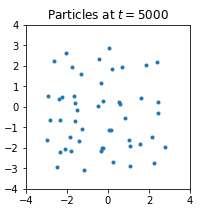}
        \caption{SGLD+R dyn.}
        \label{fig:mouse}
    \end{subfigure}
    \begin{subfigure}[b]{0.2\textwidth}
        \includegraphics[width=\textwidth]{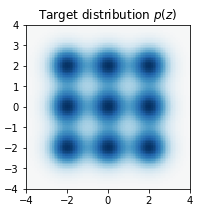}\label{fig:3x3gauss}
        \caption{MoG density}
        \label{fig:mouse}
    \end{subfigure}
    \caption{Evolution of the particles during the MoG experiment}\label{fig:mog}
\end{figure}

\begin{itemize}
\setlength\itemsep{-0.2em}
\item \textbf{Mixture of Exponentials (MoE)}. Two exponential distributions with different scale parameters $\lambda_1 = 1.5, \lambda_2=0.5$ and mixture proportions $\pi_1 = 1/3, \pi_2 = 2/3$. The pdf is
$$ 
p(z) = \sum_{i=1}^2 \pi_{i}\lambda_i \exp(-\lambda_i z).
$$
The exact value of the first and second moments can be computed using the change of variables formula
\begin{equation*}
\mathbb{E} \left[z^n \right] = \sum_{i=1}^2 \pi_{i}\frac{n!}{\lambda_i^n}
\end{equation*}
with $n\in \mathbb{N}$. 
Since $z>0$, to use the proposed scheme, we reparameterize using the $\log$ function. The pdf of the transformation $y = \log(z)$ can be computed using
\begin{equation*}
p(y) = p(\log^{-1}(y))|\mbox{D} \log^{-1}(y))|.
\end{equation*}
\item \textbf{Mixture of 2D Gaussians (MoG)}. A grid of $3 \times 3$ equally distributed isotropic 2D Gaussians, see Figure \ref{fig:mog}(d) for the density plot.  We set $\Sigma = \mbox{diag} (0.1, 0.1)$ and place the nine Gaussians centered at the following points:
\begin{align*}
\lbrace (-2,-2), (-2, 0), (-2, 2), (0, -2), \\(0, 0), (0, 2), (2, -2), (2, 0), (2, 2)  \rbrace.
\end{align*}
\end{itemize}

\begin{figure}[!ht]
    \centering
    \begin{subfigure}[b]{0.45\textwidth}
        \includegraphics[width=\textwidth]{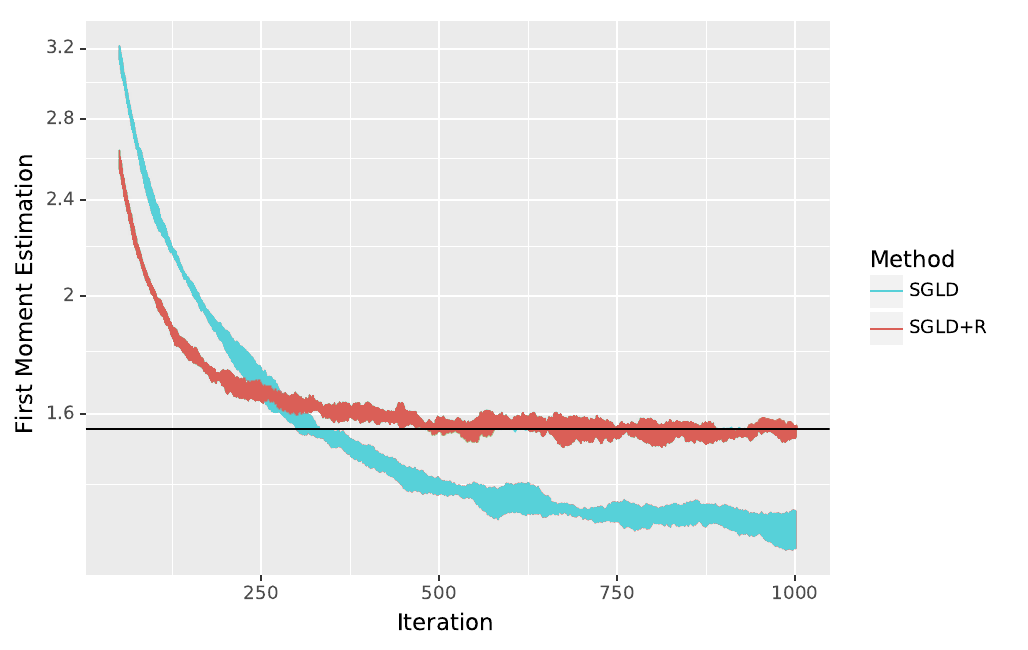}
        \caption{Estimation of $\mathbb{E}\left[X\right]$}
        \label{fig:gull}
    \end{subfigure}
    \begin{subfigure}[b]{0.45\textwidth}
        \includegraphics[width=\textwidth]{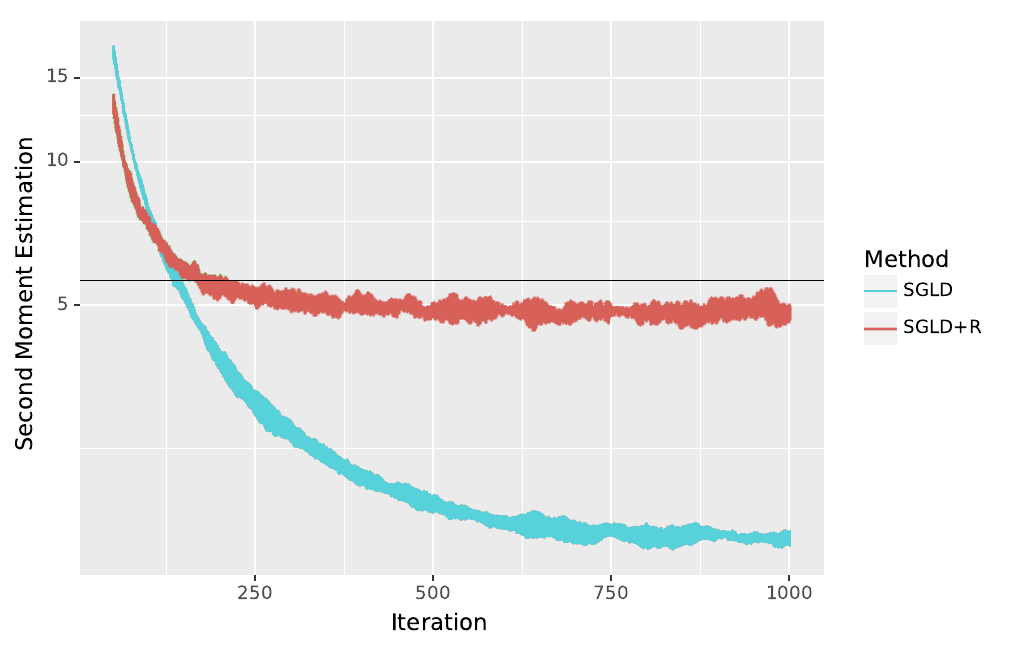}
        \caption{Estimation of $\mathbb{E}\left[X^2\right]$}
        \label{fig:tiger}
    \end{subfigure}
    \caption{Evolution of estimation during the MoE experiment. We plot the are curves for 5 simulations. 10 particles used in each simulation. Black line depicts exact value to be estimated}\label{fig:moe}
\end{figure}

\begin{figure}[!ht]
    \centering
    \begin{subfigure}[b]{0.45\textwidth}
        \includegraphics[width=\textwidth]{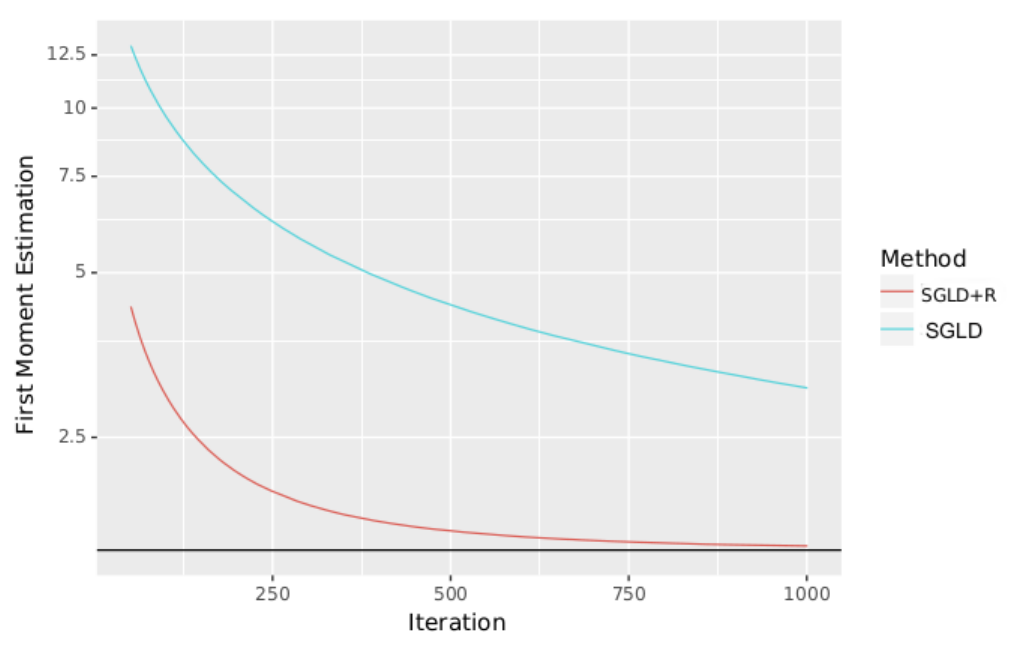}
        \caption{Estimation of $\mathbb{E}\left[X\right]$}
        \label{fig:gull}
    \end{subfigure}
    \begin{subfigure}[b]{0.45\textwidth}
        \includegraphics[width=\textwidth]{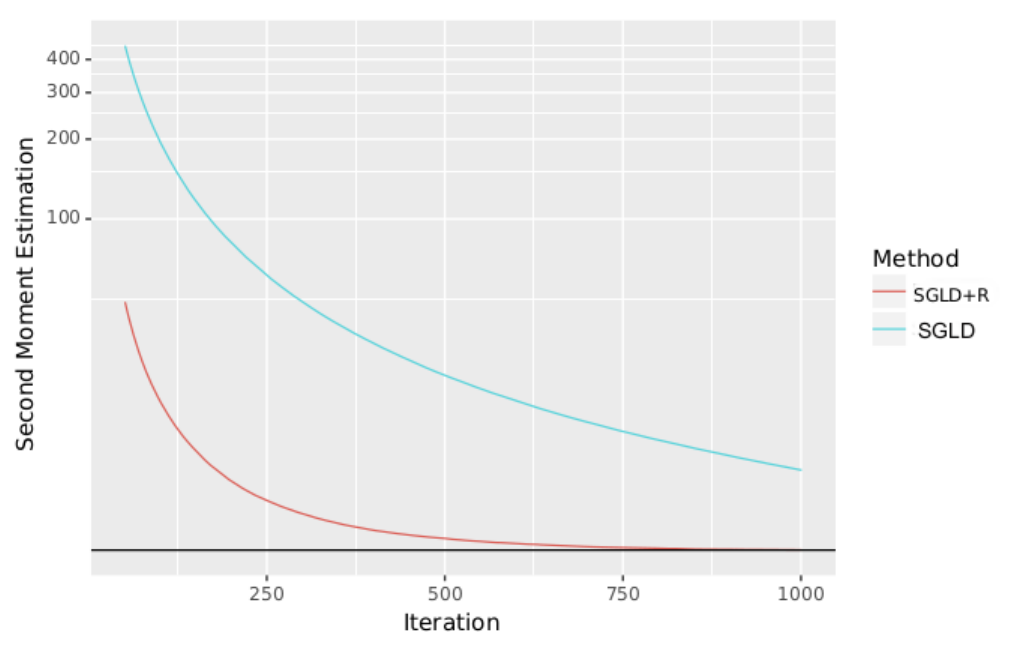}
        \caption{Estimation of $\mathbb{E}\left[X^2\right]$}
        \label{fig:tiger}
    \end{subfigure}
    \caption{Evolution of estimation during the MoE experiment. 100 particles are used and black line depicts the exact value to be estimated}\label{fig:moe100}
\end{figure}

We compare two sampling methods, SGLD with $L$ parallel chains, and our proposed scheme, SGLD+R. Note that the main difference between these two sampling algorithms is that for the former $\mathbf{\mathbf{D_K}} = \mathbf{I}$ whereas the latter accounts for repulsion between particles. Table \ref{tab:ess} reports the effective sampling size metrics, \cite{kass1998markov}, for each method using $L=10$ particles. Note that while ESS/s are similar, the repulsive forces in SGLD+R make for a more efficient exploration, resulting in much lower estimation errors. Figures \ref{fig:moe} and \ref{fig:mog} confirm this fact. In addition, even when increasing the number of particles $L$, SGLD+R achieves lower errors than SGLD (see Fig. \ref{fig:moe100}).

For the computation of the error of $\mathbb{E}\left[ X \right]$ in Tables \ref{tab:ess} and \ref{tab:ess2}, we sample for 500 iterations after discarding the first 500 iterations as burn-in, and we collect samples every 10 iterations to reduce correlation between samples. For the MoE case we used 10 particles whereas for the MoG task we used 20 particles given the bigger number of modes.

\begin{table}[H]
\centering{\small
\scalebox{0.95}{
\begin{tabular}{l|cc|cc}
\hline
& \multicolumn{2} {c} {ESS} & \multicolumn{2} {|c}{ESS/s}  \\
\textbf{Distribution} & {\bfseries SGLD} & {\bfseries SGLD+R} & {\bfseries SGLD}& {\bfseries SGLD+R}  \\
\hline
MoE& $ 44.3 $ & $ \pmb{59.1} $ & $ 51.5 $ & $ \pmb{61.0} $ \\
MoG& $151.3$ &  $ \pmb{169.5}$ &  $\pmb{36.3}$ &  $32.5$  \\
\hline
\end{tabular}
}
}
\caption{Effective sample size results for the two synthetic distributions task}\label{tab:ess}
\end{table}

\begin{table}[H]
\centering{\small
\scalebox{0.95}{
\begin{tabular}{l|cc}
\hline
& \multicolumn{2} {|c}{Error of $\mathbb{E}\left[ X \right]$} \\
\textbf{Distribution} & {\bfseries SGLD}& {\bfseries SGLD+R}  \\
\hline
MoE&  $0.39$  & $\pmb{0.14}$\\
MoG&  $1.42$ & $ \pmb{1.19} $ \\
\hline
\end{tabular}
}
}
\caption{Error results for the two synthetic distributions task}\label{tab:ess2}
\end{table}


\subsection{Bayesian Neural Network} 
We test the proposed scheme in a suite of regression tasks using a feed-forward neural network with one hidden layer with 50 units and ReLU activations. The goal of this experiment is to check that the proposed samplers scale well to real-data settings and complex models such as Bayesian neural networks. The datasets are taken from the UCI repository, \cite{Lichman:2013}. We use minibatches of size 100. 
As before, we compare SGLD and SGLD+R, reporting the average root mean squared error and log-likelihood over a test set in Table \ref{tab:bnn} and Table \ref{tab:bnn2}. We observe that SGLD+R typically outperforms SGLD. During the experiments, we noted that in order to reduce computation time, during the last half of training we could disable the repulsion between particles without incurring in performance cost.

The learning rate $\epsilon$ was chosen from a grid $\{1e-5, \ldots, 1e-3 \}$ validated on another fold. The number of iterations was set to 2000 in every experiment. As before, to make predictions we collect samples every 10 iterations after a burn-in period. 20 particles were used for each of the tested datasets.

\begin{table}[H]
\centering{\small
\scalebox{0.95}{
\begin{tabular}{l|cc}
\hline
& \multicolumn{2} {c} {Avg. Test LL}  \\
\textbf{Dataset} & {\bfseries SGLD} & {\bfseries SGLD+R}  \\
\hline
Boston&  $ -2.551\pm 0.018$ & $ -2.575 \pm 0.007$ \\
Kin8nm&  $0.826 \pm 0.005$ & $0.831 \pm 0.006$  \\
Naval&  $3.379\pm 0.011$ & $ \pmb{3.428 \pm 0.019} $  \\
Protein&   $-2.991 \pm 0.000 $ & $\pmb{-2.987 \pm 0.001} $ \\
Wine&  $ -0.765 \pm 0.008 $ & $ \pmb{-0.750 \pm 0.007}$   \\
Yacht&  $-1.211 \pm 0.020   $ & $-1.172 \pm 0.026 $  \\
\hline
\end{tabular} 
}
}
\caption{Log-Likelihood results for the BNN experiments}\label{tab:bnn}
\end{table}

\begin{table}[H]
\centering{\small
\scalebox{0.95}{
\begin{tabular}{l|cc}
\hline
& \multicolumn{2} {c} {Avg. Test RMSE}  \\
\textbf{Dataset} & {\bfseries SGLD} & {\bfseries SGLD+R}  \\
\hline
Boston& $ 2.392 \pm 0.018$ & $ \pmb{2.295 \pm 0.017}$  \\
Kin8nm& $0.104 \pm  0.001  $ & $0.104 \pm 0.001$  \\
Naval& $0.008 \pm 0.000$ & $0.008 \pm 0.000$  \\
Protein& $4.810 \pm 0.003$ & $\pmb{4.794 \pm 0.003} $  \\
Wine& $0.522 \pm 0.004$ & $\pmb{0.514 \pm 0.004}$  \\
Yacht& $0.942 \pm 0.015$ & \pmb{$0.894 \pm 0.029 $}   \\
\hline
\end{tabular} 
}
}
\caption{Root Mean Squared Error results for the BNN experiments}\label{tab:bnn2}
\end{table}

\section{CONCLUSIONS AND FURTHER WORK}\label{sec:conclusion}

This paper shows how to generate new SG-MCMC methods, such as our main contribution, SGLD+R, consisting of multiple chains plus repulsion between the particles. Instead of the naive parallelization, in which a particle from a chain is agnostic to the others, we showed how it is possible to adapt another method from the literature, SVGD, to account for a better exploration of the space, avoiding collapse between particles. We also studied the behaviour of both SVGD and SGLD+R using the Fokker-Planck equation and noting that whereas SVGD cannot be used as a valid SG-MCMC sampler, the introduction of a noise term as in SGLD+R indeed enables convergence to the actual posterior in the limit. 
Our experiments show that the proposed ideas improve efficiency when dealing 
with large scale inference and prediction problems in presence of many 
parameters and large data sets.

Several avenues are open for further work. Here we discuss only several promising ones. First, with a very large particle regime ($L >> 100$) there is room to use approximating algorithms such as Barnes-Hutt to keep the computational cost tractable.
Secondly, we used the RBF kernel in all our experiments, but a natural 
issue to address would be to define a parameterized kernel $k_{\theta} (z_i, z_j)$ and learn the parameters $\theta$ on the go to optimize the ESS/s rate, using meta-learning approaches such as the one proposed in \cite{gallego2019vis} for the SGLD case.

\vspace{1cm}

\textbf{Acknowledgments}
 VG acknowledges support from grant FPU16-05034. DRI is grateful to the MINECO MTM2017-86875-C3-1-R project and the AXA-ICMAT Chair in Adversarial Risk Analysis. All authors acknowledge support from the Severo Ochoa Excellence Programme SEV-2015-0554 and US National Science Foundation through grant DMS-1638521.




\bibliographystyle{abbrvnat} 
\bibliography{sample}

\appendix

\end{document}